\documentclass[runningheads]{llncs}
\usepackage[T1]{fontenc}
\usepackage[utf8]{inputenc}
\usepackage{multirow,tabularx}
\usepackage{ragged2e} 
\usepackage{graphicx}
\usepackage{newunicodechar}
\newunicodechar{−}{-}
\begin{document}
\title{License Plate Detection and Character Recognition Using Deep Learning and Font Evaluation
}
\author{Zahra Ebrahimi Vargoorani\inst{1}\orcidID{0009-0007-5307-3667} \and
Ching Yee Suen\inst{2}\orcidID{0000−0003−1209−7631}}
\authorrunning{Zahra E. Vargoorani, Ching Y. Suen}
\institute{Concordia University, Montreal, Quebec, Canada
\email{zahra.ebrahimivargoorani@concordia.ca}\\
\and
Concordia University, Montreal, Quebec, Canada\\
\email{chingyee.suen@concordia.ca}}
\maketitle              % typeset the header of the contribution
\begin{abstract}
License plate detection (LPD) is essential for traffic management, vehicle tracking, and law enforcement but faces challenges like variable lighting and diverse font types, impacting accuracy. Traditionally reliant on image processing and machine learning, the field is now shifting towards deep learning for its robust performance in various conditions. Current methods, however, often require tailoring to specific regional datasets. This paper proposes a dual deep learning strategy using a Faster R-CNN for detection and a CNN-RNN model with Connectionist Temporal Classification (CTC) loss and a MobileNet V3 backbone for recognition. This approach aims to improve model performance using datasets from Ontario, Quebec, California, and New York State, achieving a recall rate of 92\% on the Centre for Pattern Recognition and Machine Intelligence (CENPARMI) dataset and 90\% on the UFPR-ALPR dataset. It includes a detailed error analysis to identify the causes of false positives. Additionally, the research examines the role of font features in license plate (LP) recognition, analyzing fonts like Driver Gothic, Dreadnought, California Clarendon, and Zurich Extra Condensed with the OpenALPR system. It discovers significant performance discrepancies influenced by font characteristics, offering insights for future LPD system enhancements.

\end{abstract}

\section{Introduction}
Automatic License Plate Recognition (ALPR) systems face challenges due to environmental factors like light conditions, rain and dust, high speeds, varied angles, and poor image quality~\cite{ref_one}. The font style on license plates also significantly affects recognition~\cite{ref_two}. ALPR is crucial for traffic control, security enhancement, and law enforcement. Despite various techniques, including image processing and machine learning, there is still potential to improve accuracy in ALPR systems.

Faster R-CNN~\cite{ref_3} is a notable computer vision model that improves object detection by integrating a Region Proposal Network (RPN) with a detection network. This model increases detection speed and accuracy through shared computation across the entire image. It works in two stages: first, the RPN generates region proposals~\cite{ref_four}, which are then classified and refined by the detection network. Known for its end-to-end training, high accuracy, and fast processing, Faster R-CNN is versatile in surveillance and autonomous navigation applications, effectively handling various object sizes and scenarios.

Deep neural networks, particularly Convolutional Neural Networks (CNNs) and Recurrent Neural Networks (RNNs), have been harnessed in image and natural language processing, achieving significant improvements in object detection via deep learning. The integration Of these networks with Connectionist Temporal Classification (CTC) Loss~\cite{ref_three} has notably enhanced the accuracy and efficiency of these systems. CNNs, with their ability to handle spatial data processing, and RNNs, which excel in sequence prediction, are particularly suitable for complex scenarios. CTC Loss aligns input sequences with outputs, thereby improving traditional methods and expanding applications in automation and intelligence, such as autonomous driving and surveillance.

Typography plays a crucial role in text presentation and recognition systems. Specific typefaces designed for unique applications, like mathematical notation or music scores, enhance text presentation and comprehension. The font choice impacts the accuracy and efficiency of computer-based text recognition systems, significantly influencing their reliability and operational success.

The main contributions of this work are:
\begin{enumerate}
\item Introducing a new dataset(CENPARMI) of license plate images collected from Quebec and Ontario provinces in Canada, California, and the State of New York in the United States of America.
\item Proposing a new two-stage solution for the license plate detection phase using a Faster-RCNN model and CNN+RNN Model with Connectionist Temporal Classification(CTC) Loss networks for recognizing letters.
\item Perform an analytical evaluation of five fonts commonly used on license plates and explore how typeface design can impact computer-based studies in this context.
\end{enumerate}

\subsection{Datasets and Fonts}
\subsubsection{UFPR- ALPR Dataset:}
The UFPR-ALPR~\cite{ref_five} dataset, developed by the Federal University of Paraná, includes 4,500 diverse images of license plates from Paraná, Brazil, captured using GoPro Hero4 Silver, Huawei P9 Lite, and iPhone 7 Plus cameras. The dataset is designed for machine learning and divided into 40\% training, 20\% validation, and 40\% testing splits. It features various real-world conditions, such as different distances, sunlight, and shadows, to improve the robustness of automatic license plate recognition systems. Although generally clear, the mandatory font used on Brazilian license plates can still be confusing in specific contexts.
Figure 1 highlights potential confusion with the Mandatory font used on license plates, illustrating how '8' might be mistaken for 'B,' and '4' could be confused with 'A' and 'O' with number '0'. These examples emphasize the need to carefully evaluate the fonts on license plates to ensure clarity and avoid misinterpretation in both digital and real-world settings.
\begin{figure}[ht] 
  \label{ fig7} 
  \begin{minipage}[b]{0.4\linewidth}
    \centering
    \includegraphics[width=.5\linewidth]{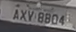} 
    \vspace{1ex}
  \end{minipage}%%
  \begin{minipage}[b]{0.4\linewidth}
    \centering
    \includegraphics[width=.5\linewidth]{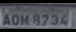} 
    \vspace{1ex}
  \end{minipage} 
  \centering
      \caption{Confusing glyphs of Mandatory font in license plate context}\label{fig3}
\end{figure}
\subsubsection{CENPARMI Dataset:}
The Centre for Pattern Recognition and Machine Intelligence (CENPARMI) has developed a dataset containing 1,600 license plates from diverse regions, including California, New York, Ontario, and Quebec, captured using a mobile phone camera. This dataset addresses challenging lighting conditions such as harsh sunlight, shadows, and dimly lit environments, significantly affecting license plate visibility. It includes plates photographed from 1 to 30 meters and features about 20\% of the plates shot at angles, adding complexity due to their non-standard orientations. This diverse collection is a valuable tool for advancing license plate recognition technologies.

Moreover, the dataset is not just a theoretical exercise, but a practical resource that encompasses a range of typefaces used on license plates, including Driver Gothic for Quebec and Dreadnought for Ontario, Clarendon for California, and Zurich Extra Condensed for New York. It also spans multiple vehicle types—motorcycles, cars, and public transport vehicles—and illustrates variations in image brightness, vehicle distance, plate sizes, and background colors, all contributing to the complexities faced by recognition systems. This diversity ensures the dataset's applicability in real-world scenarios, making it a critical resource for developing robust recognition technologies.

The paper includes a review of related literature, introduces a proposed model, discusses the results, and concludes with future directions in license plate recognition technology.
\begin{figure}[ht] 
  \label{ fig7} 
  \begin{minipage}[b]{0.2\linewidth}
    \centering
    \includegraphics[width=.8\linewidth]{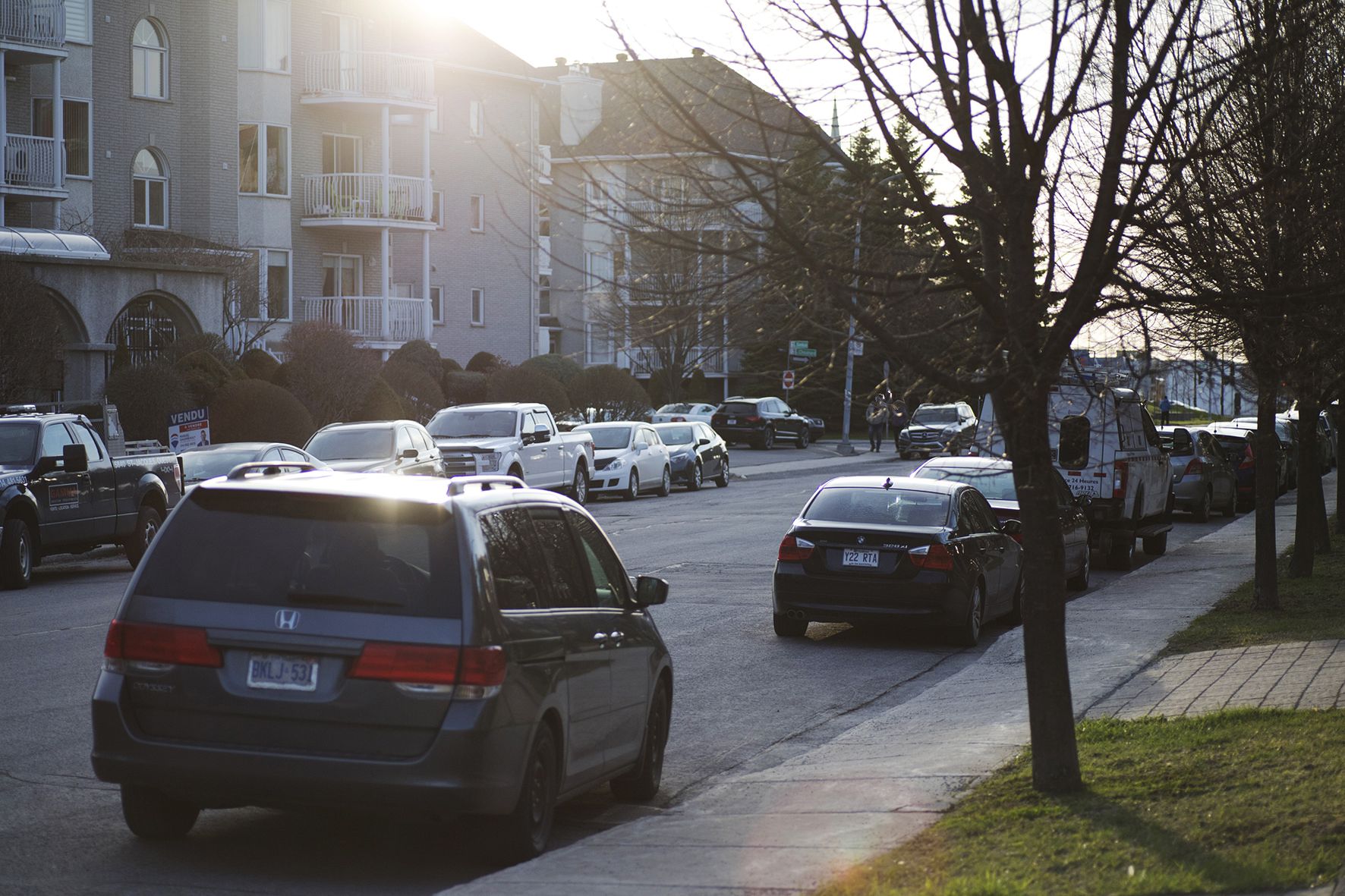} 
    \vspace{3ex}
  \end{minipage}%%
  \begin{minipage}[b]{0.2\linewidth}
    \centering
    \includegraphics[width=.8\linewidth]{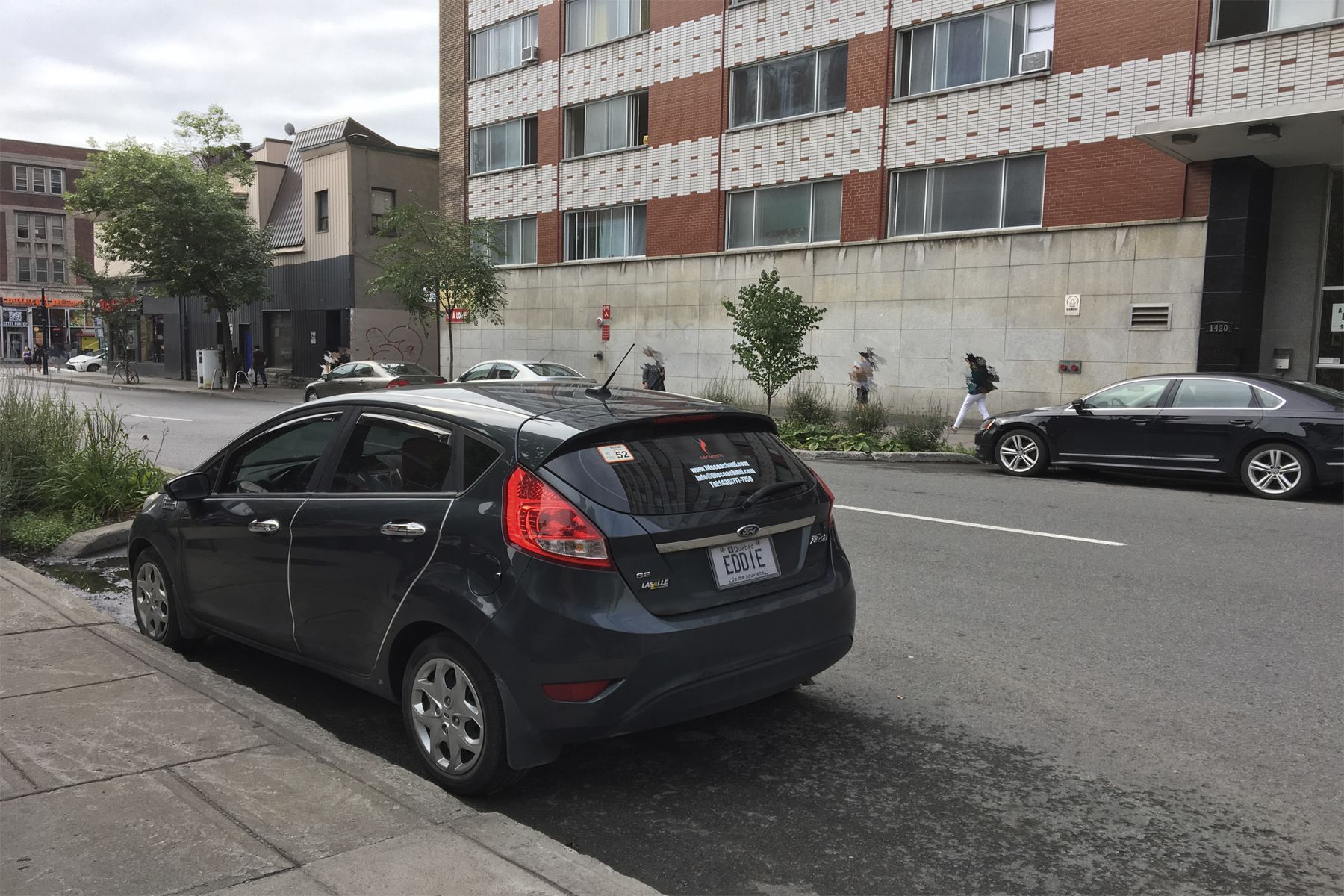} 
    \vspace{3ex}
  \end{minipage} 
  \begin{minipage}[b]{0.2\linewidth}
    \centering
    \includegraphics[width=.8\linewidth]{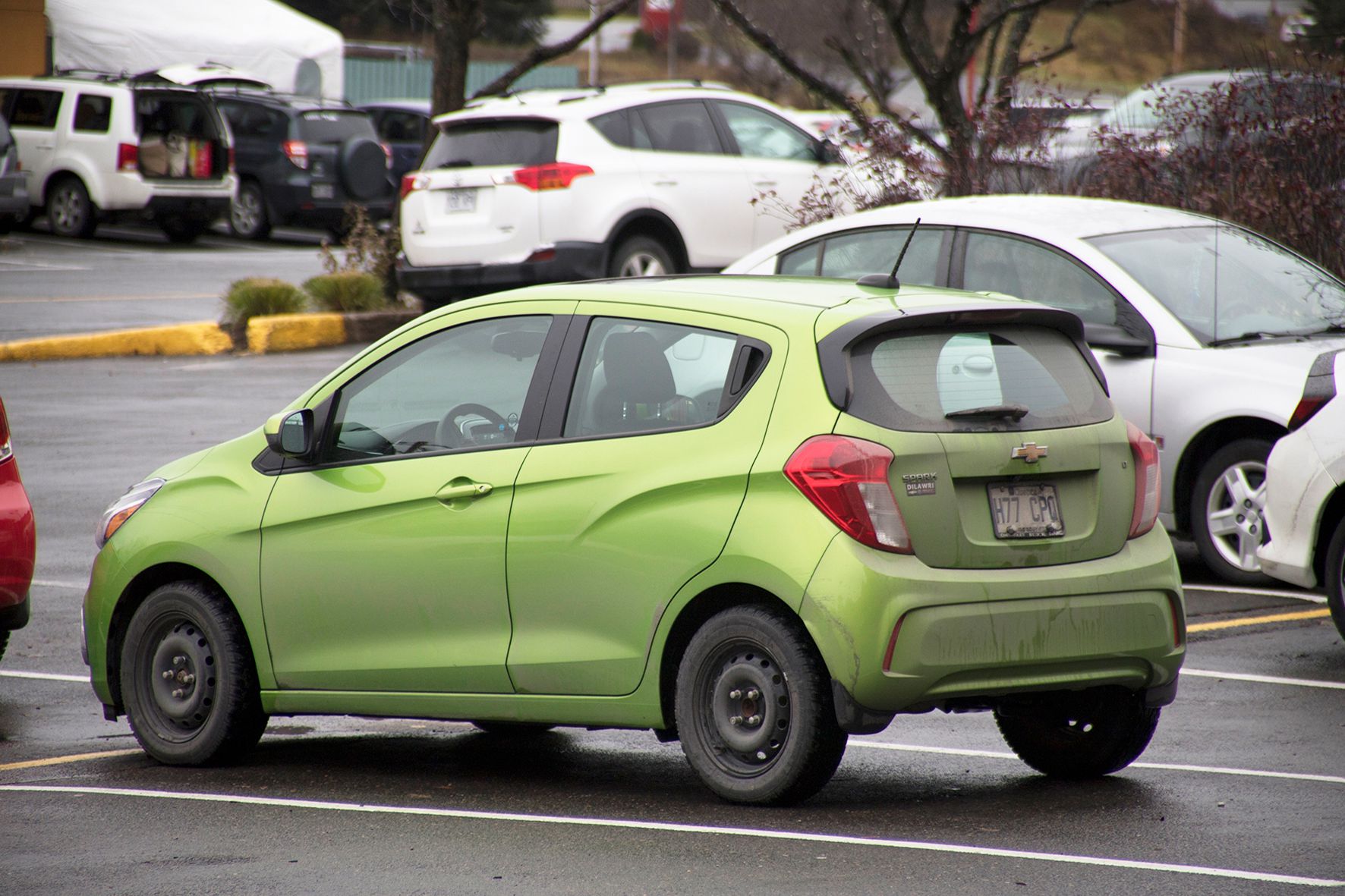} 
    \vspace{3ex}
  \end{minipage}%% 
  \begin{minipage}[b]{0.2\linewidth}
    \centering
    \includegraphics[width=.8\linewidth]{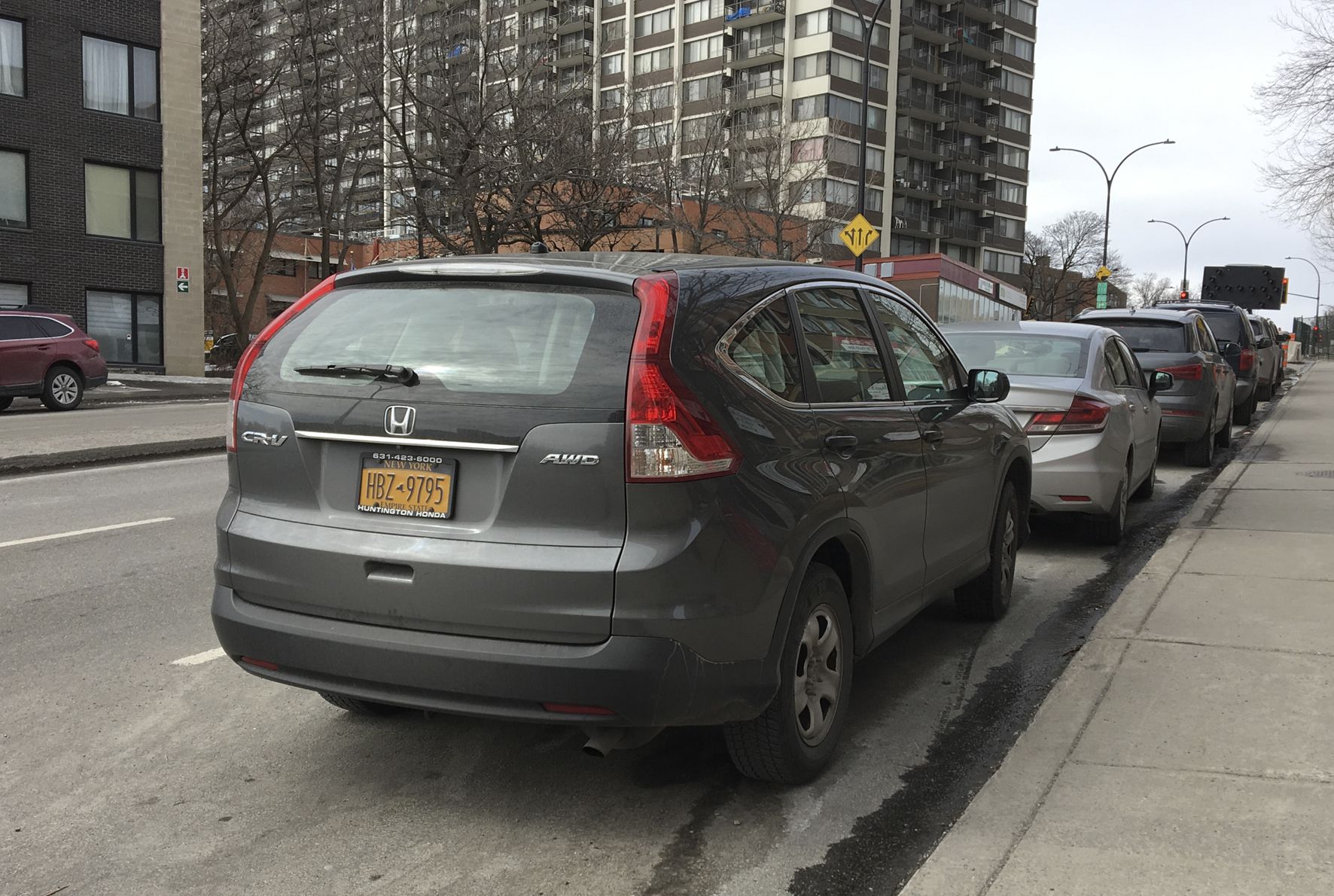} 
    \vspace{3ex}
  \end{minipage} 
  \centering
      \caption{Sample images from CENPARMI Dataset
}\label{F1}
\end{figure}
\section{Related Work}

In reference \cite{ref_six}, researchers used a Convolutional Neural Network (CNN) to extract relevant features from license plates, identifying distinct visual patterns through convolutional layers. These features were then processed by a Recurrent Neural Network (RNN) with 36 hidden units to sequence the detected characters and accurately reconstruct the license plate information. By combining the spatial pattern recognition capabilities of CNNs with the sequential learning abilities of RNNs, they achieved a 76\% accuracy rate for recognizing all characters on a license plate and a 95.1\% accuracy rate per individual character. This approach highlights the complementary strengths of CNNs and RNNs: CNNs capture intricate visual details, while RNNs process sequences of characters in the correct order.

In reference~\cite{ref_eight}, the Single Shot Detection (SSD) algorithm, coupled with Residual Network (ResNet) instead of the typical VGG Network, was used for license plate detection. ResNet helped prevent gradient vanishing, enhancing performance and achieving an 85.5\% accuracy rate, better than VGG's 83.6\%. This study emphasizes the importance of refining algorithms to improve effectiveness in computer vision tasks.

Reference~\cite{ref_nine} described modifications to the FAST-YOLO network for detecting vehicles and their license plates simultaneously, adapting YOLO for character detection and integrating heuristic strategies. This resulted in a 63.18\% accuracy rate in detecting and recognizing license plates, surpassing the 55.47\% accuracy of the Sighthound method. This improvement underscores the benefits of customizing neural network architectures and using heuristic techniques to enhance precision in computer vision tasks.

The research in~\cite{ref_ten} highlighted the impact of font characteristics on word recognition and reading comprehension in digital publishing. It introduced a new shape descriptor for recognizing ancient characters with high accuracy, accommodating variations like translation, rotation, and scaling. This study underlined the relationship between typography and readability, showing how design choices enhance digital reading experiences.

The study in~\cite{ref_eleven} outlined the development of Sitka, a new serif typeface designed to enhance font legibility on digital displays. This typeface was part of a broader initiative aimed at facilitating legibility studies and iterative design processes, emphasizing trade-offs in design, especially regarding x-height, which affects the legibility of letters with ascenders and descenders. This research advances typeface design, highlighting challenges in optimizing legibility for narrow letters.

\section{Proposed Methodology}
While this work leverages established approaches in the field of license plate detection and recognition, it enhances them through strategic combinations and novel applications to achieve improved performance. The methodological framework consists of the following key components:

\subsection{License Plate Detection}
The license plate detection is performed using a Faster R-CNN~\cite{ref_3} (Region-based Convolutional Neural Network) architecture, which is widely recognized for its robustness and high detection accuracy, even in challenging conditions such as varying lighting, occlusions, and diverse plate designs. Our approach utilizes a ResNet-50 backbone pre-trained on ImageNet to balance accuracy and computational efficiency while extracting high-level features from input images. The Region Proposal Network (RPN) generates region proposals likely to contain license plates, refining and filtering these proposals to reduce false positives. These proposals undergo ROI (Region of Interest) pooling and are passed through fully connected layers for precise bounding box regression and classification.

\subsection{License Plate Recognition}
For the OCR task, we developed a CNN-RNN model integrated with Connectionist Temporal Classification(CTC) loss~\cite{ref_three}. This combination is particularly effective for sequence-based recognition tasks such as reading the characters on a license plate. The CNN feature extractor, inspired by MobileNetV3, is lightweight yet powerful, designed to run efficiently on edge devices while extracting spatial features from the detected license plate images. The RNN sequence modeler, using a bidirectional LSTM (Long Short-Term Memory) network, processes the sequential features extracted by the CNN, capturing contextual information across the character sequence. The Connectionist Temporal Classification(CTC) loss allows the model to be trained end-to-end for sequence prediction without requiring pre-segmented training data, crucial for handling varying lengths and unaligned character sequences typical of license plates.

\subsection{Model Training and Optimization}
The recognition model underwent 100 iterations or epochs during the training phase to ensure optimal convergence and proficiency. We chose to use 100 iterations for training the model based on empirical observations. This number of iterations allowed for optimal convergence and proficiency without overfitting the model, balancing training time and performance. The learning rate was set to 0.01, determining the step size at which the stochastic gradient descent optimizer adjusted the model's parameters during the back-propagation process. This choice of learning rate was informed by empirical observations and experimentation, balancing the need for stable convergence and fine-grained parameter updates. To train the model's ROI (Region of Interest) heads, we utilized a batch size of 64. This batch size indicates the number of ROI samples per image used in the training process. By carefully selecting this batch size, we aimed to balance computational efficiency and the model's ability to generalize to varying regions of interest within license plate images.

In our study, OpenALPR serves as a baseline for comparison. This widely recognized system, developed by OpenALPR Technology Inc. and launched in May 2014, can detect license plates from countries such as Canada, Brazil, the USA, Japan, and China. By comparing our proposed model's performance against OpenALPR~\cite{ref_tweleve}, we aim to highlight the improvements and advantages of our approach. While OpenALPR provides a general solution, our model demonstrates enhanced performance, particularly under challenging conditions and with diverse font styles, as evidenced by higher recognition and lower character error rates.

Three sets of results will be obtained. The initial set will present the Average Precision (AP) values for the detection task, utilizing Faster-RCNN~\cite{ref_3}. The second set will detail the mAP(mean Average Precision) values for the recognition tasks implemented with Connectionist Temporal Classification(CTC) networks~\cite{ref_three}. The final set will include qualitative results, showcasing the confusion matrices for each dataset, which will aid in determining the readability of one font over another. We recommend enhancing automatic license plate recognition systems and font features based on these findings.

\subsection{Data Augmentation and Synthetic Data Generation}
To enhance the robustness of the recognition model, various data augmentation techniques were employed. These included rotation, perspective transform, color channel transform, and adding noise. Rotation was applied in the range of -15 to 15 degrees to simulate different viewing angles. Perspective transforms were used to mimic various distortions. Color channel transformations involved adjustments to hue (hsv\_h: 0.015), saturation (hsv\_s: 0.7), and value (hsv\_v: 0.4) to account for different lighting conditions. Additionally, translations (translate: 0.1) and scaling (scale: 0.5) were applied to simulate different image perspectives, and a mosaic effect (mosaic: 1.0) was added to increase variability. These techniques significantly increased the dataset’s variability, allowing the model to learn from a broader range of examples. The data augmentation and synthetic data generation contributed to improved model performance, particularly in handling challenging conditions such as occlusions, different lighting, and varied plate designs.

\subsection{Font Type Evaluation Aspects}

Our research evaluates font types based on several key characteristics from reference~\cite{ref_two}: similar apexes, defined as the junction of a stem; positioning of crossbars, which are horizontal strokes connecting parts of a letter, with varied positions enhancing legibility; similar top and bottom counters, the open spaces within a letter, with identical bottom counters increasing the likelihood of confusion; similar bowls, the rounded parts of letters; identical spurs, small projections at the end of a curved stroke; identical horizontal strokes at the bottom or top of letters; unclear tails, short downward strokes that can be mistaken for other elements when not clear; and similar diagonal strokes, where identical placement and design can lead to confusion. These criteria were chosen because they directly impact the readability and recognizability of different fonts. By analyzing these aspects, we aim to determine which font features contribute to better recognition performance and why certain fonts may lead to higher error rates.

\section{Results and Discussion}

For the training process of detection and recognition models, we leveraged the power of the NVIDIA Tesla T4 GPU. The NVIDIA Tesla T4 has 16GB of GPU memory, offering substantial computational resources and parallel processing capabilities. This GPU enabled us to accelerate the training and optimization of our models, effectively handling large-scale datasets.

\subsection{Evaluation Metrics}
The performance of our models on the test set was evaluated using various metrics. The average precision (AP) at different IOU thresholds, areas, and maximum detections provide insights into the model’s object detection capabilities. For the recognition model, we used recognition rate, character error rate (CER) and recall ratio as evaluation metrics. Additionally, confusion matrices were computed to analyze the types of errors made by the recognition model.

\subsection{Detection Results}
The detection model maintained a high precision at a higher IoU threshold of 0.75. When considering different object sizes, the model exhibited varying performance. For small objects, the average precision needed to be computed (-1.000), indicating a lack of reliable detection. However, for medium-sized objects, the model achieved a reliable average precision. The model excelled in detecting large objects, achieving an Impressive average precision. These results demonstrate the effectiveness of our Faster R-CNN~\cite{ref_3} model in accurately detecting license plates in various scenarios, highlighting its potential for real-world license plate recognition applications(see Fig. 3). The results of the detection models for four datasets are shown in Table 1.

\begin{figure}[ht] 
  \label{ fig7} 
  \begin{minipage}[b]{0.25\linewidth}
    \centering
    \includegraphics[width=.8\linewidth]{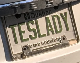} 
    \vspace{4ex}
  \end{minipage}%%
  \begin{minipage}[b]{0.25\linewidth}
    \centering
    \includegraphics[width=.8\linewidth]{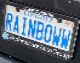} 
    \vspace{4ex}
  \end{minipage} 
  \begin{minipage}[b]{0.25\linewidth}
    \centering
    \includegraphics[width=.8\linewidth]{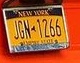} 
    \vspace{4ex}
  \end{minipage}%% 
  \begin{minipage}[b]{0.25\linewidth}
    \centering
    \includegraphics[width=.8\linewidth]{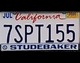} 
    \vspace{4ex}
  \end{minipage} 
      \caption{Detection sample results from left to right for Quebec and Ontario provinces and States of New York and California}\label{fig3}
\end{figure}

\begin{table}[h!]
\centering
\begin{tabular*}{\textwidth}{@{\extracolsep{\fill}} lcc}
\hline
Dataset    & AP50:95      & AP75   \\ \hline
Quebec  & 79.54 \% & 94.16 \%  \\
Ontario    & 80.37  \%  & 94.95 \%  \\
New York & 78 \% & 92.14 \%  \\
California     & 83.87 \% &  95.72 \% \\
\end{tabular*}
\vspace{2ex}
\caption{AP50:95, and AP75 scores for various datasets}
\label{table:ap_scores}
\end{table}
\subsection{Recognition Results}
The recognition model was evaluated on two datasets: the UFPR-ALPR~\cite{ref_five} dataset and our CENPARMI dataset. The results indicate a significant improvement in performance when using the CENPARMI dataset, attributed to the extensive data augmentation and synthetic data generation techniques employed during training. The results are summarized in the Table 2.

The higher recognition rate and lower CER on the CENPARMI dataset suggest that the model generalizes well to the varied conditions and plate designs encountered in real-world scenarios, and the effectiveness of our advanced training techniques. This improvement is particularly noteworthy given the diversity of the CENPARMI dataset, which includes images from different states with varying plate designs and used fonts, lighting conditions, and environmental factors.

One of the key innovations in our approach is the use of a lightweight backbone inspired by MobileNetV3 for the recognition model. This design choice ensures that the model can run efficiently on edge devices with limited computational resources. The lightweight architecture reduces the model's complexity and computational cost without compromising accuracy, making it suitable for real-time applications on devices such as smartphones, tablets, and embedded systems.
\begin{table}[h!]
\centering
\begin{tabular*}{\textwidth}{@{\extracolsep{\fill}} lcccc}
\hline
Model        & Dataset        & Recognition Rate & Character Error Rate (CER) & Recall \\ \hline
\multirow{2}{*}{our model} & UFPR-ALPR      & 84.2 \%     & 5.3 \%                         & 92.1\% \\
                           & CENPARMI  & 89.8 \%     & 4.5 \%                         & 90.0 \% \\ \hline
\multirow{2}{*}{open alpr} & UFPR-ALPR      & 71.7 \%     & 15.5 \%                         & 55.8\% \\
                           & CENPARMI  & 65.5 \%     & 19.6 \%                         & 80.2 \% \\ \hline
\end{tabular*}
\vspace{1ex}
\caption{Performance metrics for different models and datasets}
\label{table:model_metrics}
\end{table}
\subsection{State-Wise Recognition Performance}
The recognition model's performance was further analyzed on datasets from four different datasets in Canada and United States of America. The results for each dataset are detailed in the Table 3.

\begin{table}[h!]
\centering
\begin{tabular*}{\textwidth}{@{\extracolsep{\fill}} lccc}
\hline
State & Recognition Rate & Character Error Rate (CER) & Recall \\ \hline
Quebec         & 91.6\%               & 3.8 \%                                  & 91.5 \%             \\
Ontario         & 91.6\%               & 3.8 \%                               & 91.5 \%             \\
New York         & 80.1\%               & 8.1 \%                               & 80.6 \%             \\
California        & 96\%               & 2.3 \%                                  & 96.4 \%             \\ \hline
\end{tabular*}
\vspace{1ex}
\caption{Performance metrics for different states}
\label{table:state_metrics}
\end{table}

The model demonstrated consistent performance across different dataset, with slight variations attributed to differences in plate designs, font styles, and environmental conditions. California State achieved the highest accuracy and the lowest CER, likely due to clearer and more consistent plate designs. State of New York also performed exceptionally well, indicating that the model can handle diverse plate designs with high accuracy.

% \begin{figure}
% \includegraphics[width=1\linewidth]{recog_result.jpg}
% \caption{Recognition Sample Results} \label{fig2}
% \end{figure}

\begin{figure}[ht] 
  \begin{minipage}[b]{0.5\linewidth}
    \centering
    \includegraphics[width=.8\linewidth]{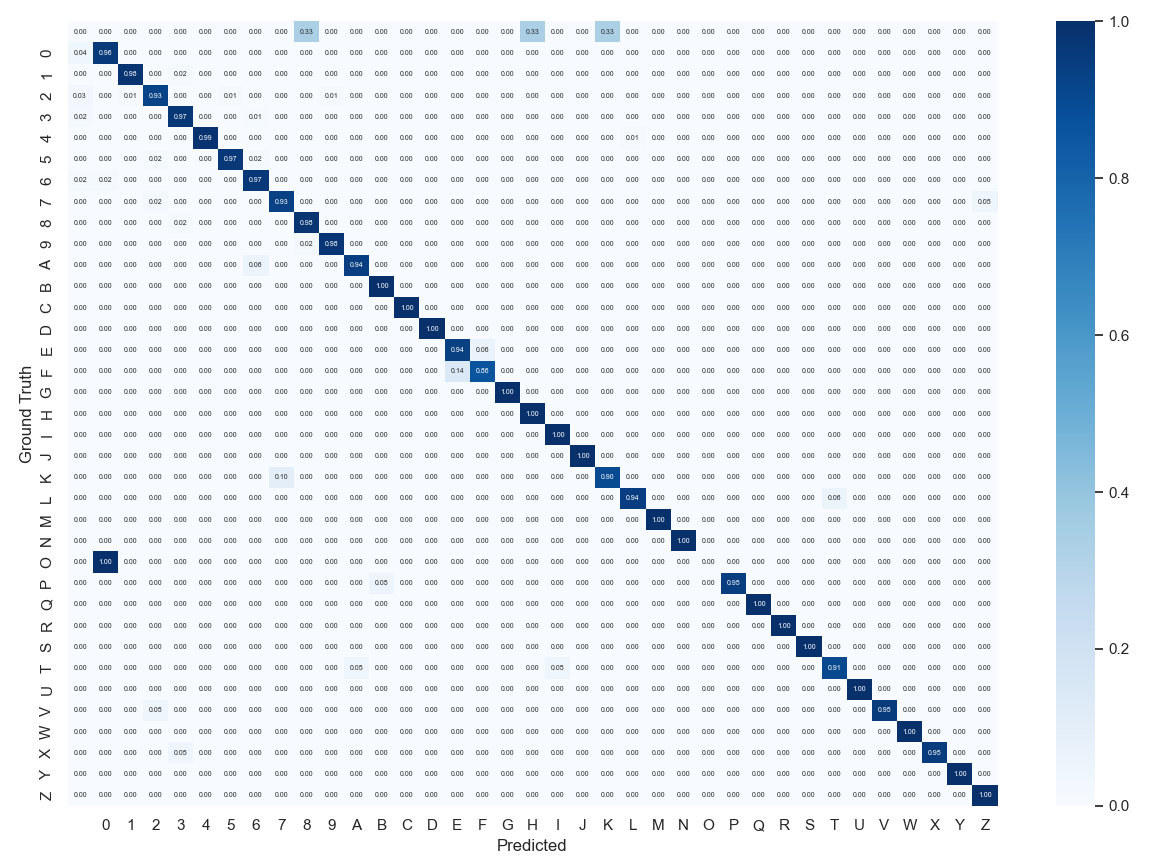} 
    \caption{Confusion matrix for California} 
    \label{fig:ca}
    \vspace{4ex}
  \end{minipage}%%
  \begin{minipage}[b]{0.5\linewidth}
    \centering
    \includegraphics[width=.8\linewidth]{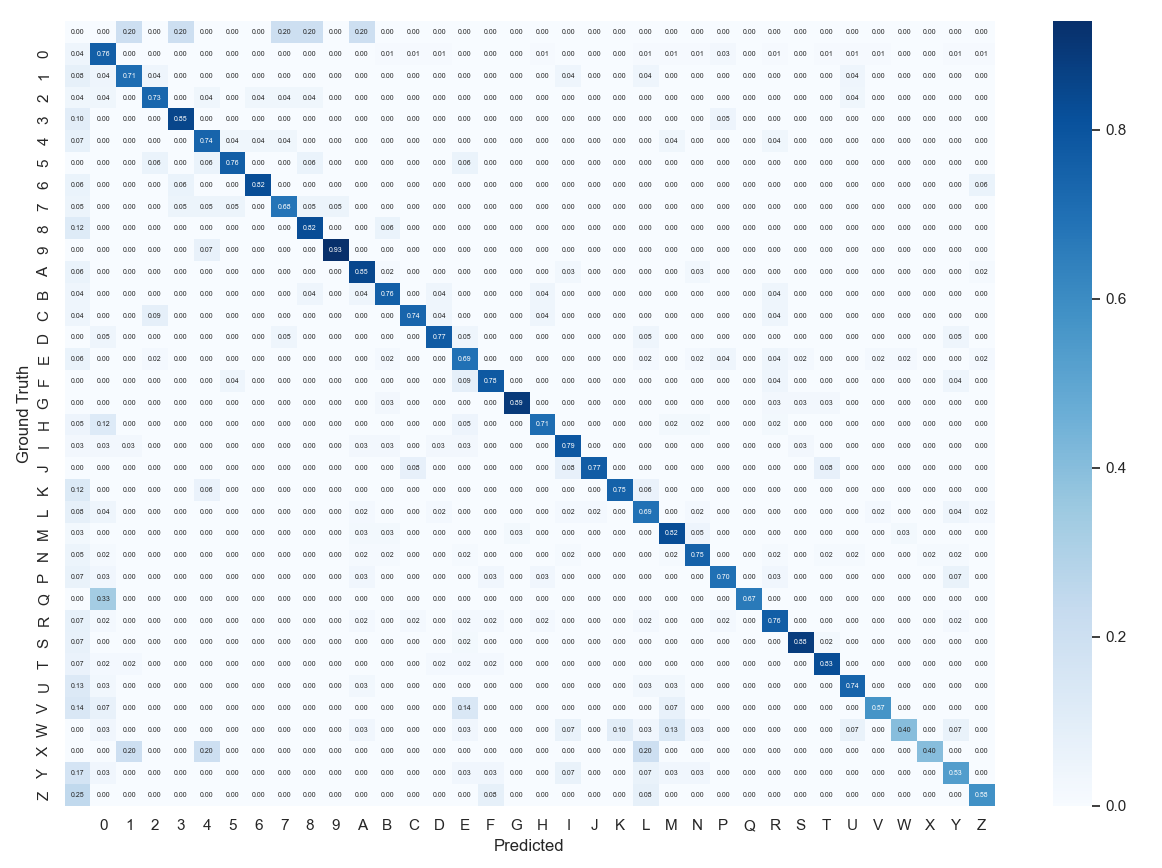} 
    \caption{Confusion matrix for New York} 
    \label{fig:ny}
    \vspace{4ex}
  \end{minipage} 
  \begin{minipage}[b]{0.5\linewidth}
    \centering
    \includegraphics[width=.8\linewidth]{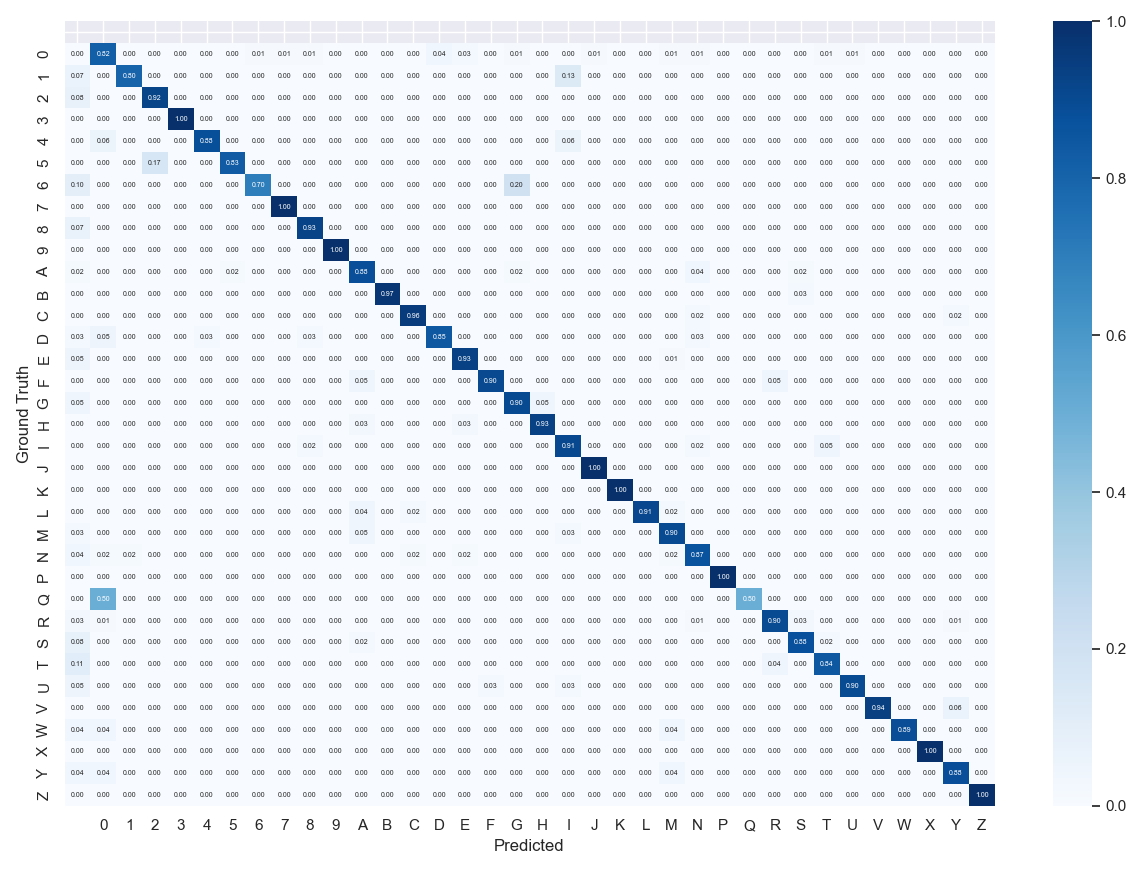} 
    \caption{Confusion matrix for Ontario} 
    \label{fig:on}
    \vspace{4ex}
  \end{minipage}%% 
  \begin{minipage}[b]{0.5\linewidth}
    \centering
    \includegraphics[width=.8\linewidth]{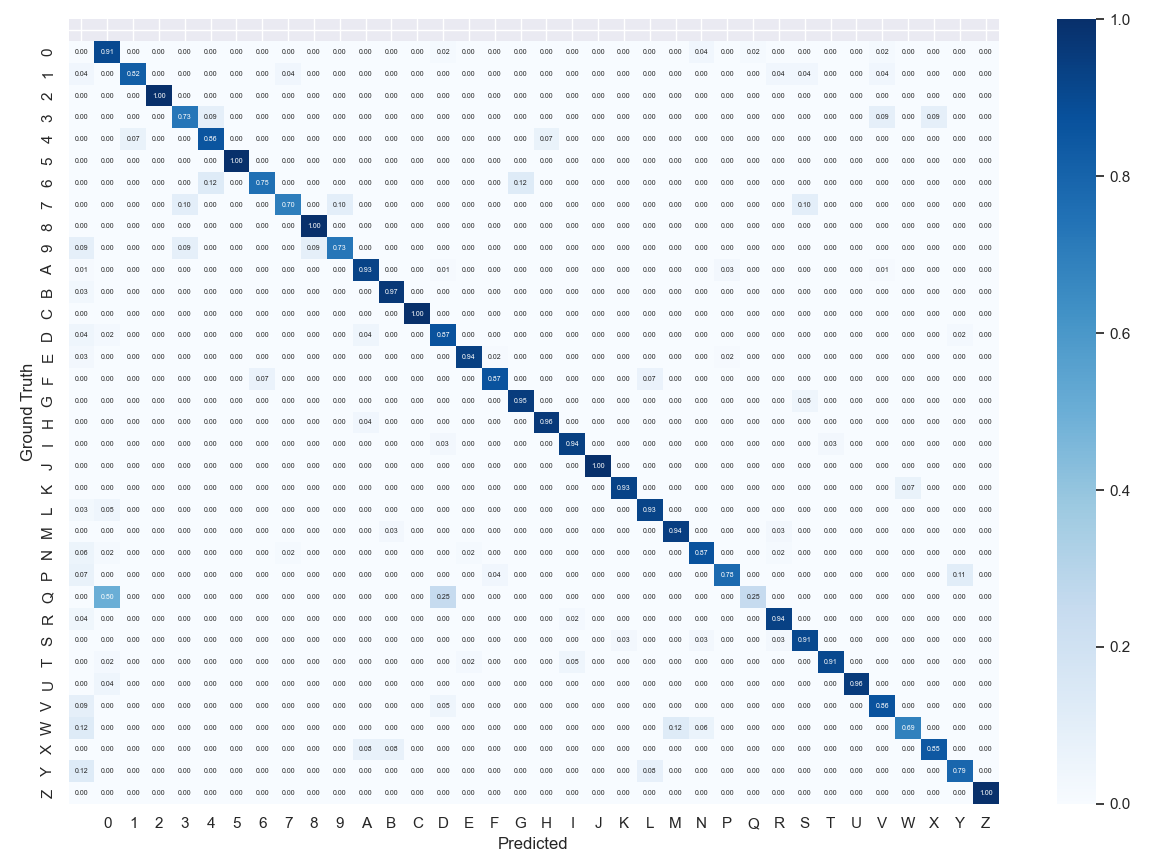} 
    \caption{Confusion matrix for Quebec} 
    \label{fig:qc}
    \vspace{4ex}
  \end{minipage}
  \caption{Confusion matrix of our proposed model on the CENPARMI and UFPR-ALPR dataset}
  \label{fig:all_states}
\end{figure}

\subsection{Font Analysis Results}
In this section, we present five distinct sets of outcomes from evaluating fonts based on the recognition results of individual letters(see Fig. 8). From our confusion matrices and  detailed analysis of the task of recognizing license plates, we can draw several conclusions:
\subsubsection{CENPARMI Dataset:} Our CENPARMI dataset includes 1600 license plates from diverse California, New York, Ontario, and Quebec environments. We conducted license plate detection and recognition in each province separately because each dataset has a unique font.
\begin{enumerate}
\item \textbf {Quebec Dataset:} The characters 'Q' and '0' often need clarification due to their similar shapes, with the 'Q's tail sometimes indistinct. The resemblance between '6' and 'G' stems from their similar curves and inner spaces. Confusion between 'Q' and 'D' can occur if the 'Q's tail blends with its bowl, resembling a 'D,' especially if both have a similar spur. 'W' and 'M' are also confusing due to their similar diagonal strokes and internal structure. Additionally, '6' and '4' can be mistaken for each other due to similarities in their upper structures and angles.
\item \textbf {Ontario Dataset:} The characters 'Q' and '0' might be mistaken for each other due to their rounded bowls, especially if the 'Q' has a subtle tail. Similarly, '5' and '2' could be confused because of their upper curves and strokes, while '6' and 'G' might look alike due to their shape and internal space. Characters like '1' and 'I', as well as '4' and 'A', can be indistinguishable when depicted as simple vertical lines. 'V' and 'Y' can also appear similar if the 'Y' has a short or blended tail.
\item\textbf  {California Dataset:} The characters 'O' and '0' can easily be confused due to their similar rounded shapes. Similarly, 'F' and 'E' might be mistaken for each other because of their horizontal strokes, making it hard to notice the missing middle stroke in 'F'. 'L' and 'T' can also appear similar due to their vertical and horizontal line components, especially in bold weights. Additionally, '7' and 'Z' might be confused when '7' has a pronounced horizontal stroke and 'Z' is styled with straight, minimalistic angles.
\item\textbf  {New York Dataset:} 'O' and '0' can look very similar due to their tightly rounded shapes. The letters 'J,' 'I,' and 'T' might be mistaken for one another due to similar vertical strokes, especially when 'J' has a subtle curve and 'T' has a short horizontal stroke. 'V' and 'U' can also be confused if 'V' has a sharp vertex resembling a 'U' without the middle crossbar. 'X' and 'L' may appear similar if 'X's diagonal strokes blend together, looking like intersecting 'L's. In a condensed font, 'M' and 'W' might look like mirror images with unclear middle peaks and valleys.
\end{enumerate}
\subsubsection{UFPR-ALPR Dataset:} The confusion matrix highlights common character misidentifications due to their similar appearances(see Fig. 9). 'O' and '0' are often confused as both have round, unmarked shapes. 'I', '1', and 'T' can be mistaken for each other when depicted as straight lines, especially if 'T' has a short horizontal stroke. 'M' and 'W' look alike due to their mirrored structures and bold lines. 'B' and '8', as well as 'E' and '8', are confusing when 'B' and 'E' resemble the double loops of '8'. 'C' and 'G' might be confused if 'G's tail is subtle, making it look like 'C'.

\begin{figure}[ht] 
  \begin{minipage}[b]{0.9\linewidth}
    \centering
    \includegraphics[width=.7\linewidth]{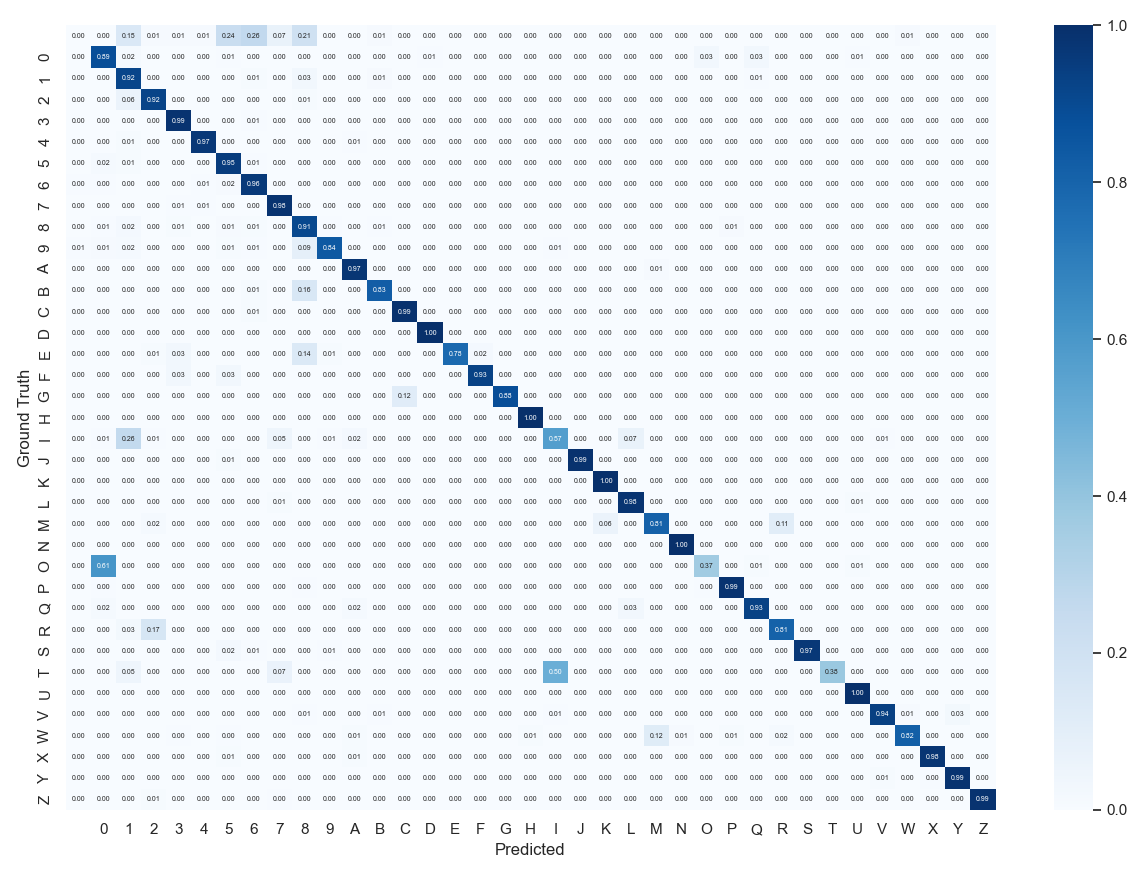} 
    \label{fig:qc}
    \vspace{1ex}
  \end{minipage}
  \caption{Confusion matrix of our proposed model on the UFPR-ALPR dataset}
  \label{fig:all_states}
\end{figure}

To better understand the results, we conducted a detailed statistical analysis to evaluate the performance of our model.
We compared our approach with the OpenALPR baseline. The results highlight our method's strengths and demonstrate its superiority in terms of recognition rate and character error rate.
We conducted a detailed error analysis to identify the causes of false positives and negatives. Common errors included misclassification due to similar font characteristics and issues with plate visibility under extreme lighting conditions.
Our experimental results demonstrate the robustness and accuracy of the proposed method across diverse conditions. While our approach builds on established techniques, the strategic combination and novel application of these methods significantly advance the field of license plate recognition, sparking new interest and possibilities. The statistical and comparative analyses further validate the effectiveness of our approach, providing strong evidence for its superiority over existing methods.

\section{Conclusion and Future Work}
In summary, this research concentrates on detecting license plates and recognizing their characters through a dual-phase approach that utilizes a Region-based Convolutional Neural Network (R-CNN) for detection and a Connectionist Temporal Classification(CTC) network for character recognition. It also examines how font selection affects ALPR system performance. The paper introduces a novel dataset and underscores the importance of precise LPD systems. Future work could include enlarging the dataset, employing additional deep learning methods, expanding the system to encompass license plate recognition and generation, and selecting a more legible font for license plates. This research contributes to advancing intelligent transportation systems, supporting improved traffic management and security.


\end{document}